\documentclass[conference]{IEEEtran}
\IEEEoverridecommandlockouts
\usepackage{cite}
\usepackage{amsmath,amssymb,amsfonts}
\usepackage{algorithmic}
\usepackage{graphicx}
\usepackage{textcomp}
\usepackage{xcolor}
\usepackage{amsmath}
\usepackage{hyperref}
\usepackage{bbold} 
\def\BibTeX{{\rm B\kern-.05em{\sc i\kern-.025em b}\kern-.08em
    T\kern-.1667em\lower.7ex\hbox{E}\kern-.125emX}}

\begin{document}

\title{TrustGAN: Training safe and trustworthy deep learning models
through generative adversarial networks\\
}

\author{\IEEEauthorblockN{H\'{e}lion du Mas des Bourboux}
\IEEEauthorblockA{\textit{Thales SIX Theresis} \\
1 av. Augustin Fresnel, 91120 Palaiseau, France \\
helion.dumasdesbourboux'at'thalesgroup.com}
}

\maketitle

\begin{abstract}
Deep learning models have been developed for a variety of tasks and
are deployed every day to work in real conditions.
Some of these tasks are critical and models need to be trusted and safe,
e.g. military communications or cancer diagnosis.
These models are given real data, simulated data or combination of both
and are trained to be highly predictive on them.
However, gathering enough real data or simulating them to be
representative of all the real conditions
is: costly, sometimes impossible due to confidentiality and most of
the time impossible.
Indeed, real conditions are constantly changing and sometimes are intractable.
A solution is to deploy machine learning models that are able to give
predictions when they are confident enough otherwise raise a flag
or abstain.
One issue is that standard models easily fail at detecting out-of-distribution
samples where their predictions are unreliable.

We present here TrustGAN, a generative adversarial network pipeline targeting trustness.
It is a deep learning pipeline which
improves a target model estimation of the confidence without impacting its
predictive power.
The pipeline can accept any given deep learning model which outputs a prediction
and a confidence on this prediction. Moreover, the pipeline does not need
to modify this target model.
It can thus be easily deployed in a MLOps (Machine Learning Operations) setting.

The pipeline is applied here to a target classification model
trained on MNIST data to recognise numbers based on images.
We compare such a model when trained in the standard way and with
TrustGAN. We show that on out-of-distribution samples, here FashionMNIST
and CIFAR10, the estimated confidence is largely reduced.
We observe similar conclusions for a classification model
trained on 1D radio signals from AugMod, tested on RML2016.04C.
We also publicly release the code.
\end{abstract}


\section{Introduction}

Deep learning models are being deployed for a large variety of tasks.
Even though a lot of these tasks do not primarily require decisions to be
confident, e.g. advertisement or content suggestions, regulators around the world
are expecting companies to build safer and trustable artificial intelligent (AI)
systems (e.g. the European Union proposal for AI regulation \cite{eu-52021PC0206}).
Other AI tasks, especially for critical infrastructures,
need to be robust and safe at their core before being deployed in
the real world, e.g. secured communication systems, radio surveillance,
navigation systems.
Building safe deep learning models which can be trusted in the real world is
a complex objective. One could think of gathering enough data, but that solution is either
too costly, impossible due to confidentiality issues or simply impossible.
Indeed a lot of tasks are constantly evolving, e.g. face masks 
after the \mbox{covid-19} pandemic and their effect on face recognition systems
(c.f. \cite{NISTmasks}).
One could think of modeling the data, but this is not possible in all situations.

\begin{figure}[ht]
    \centering
    \includegraphics[width=0.99\linewidth]{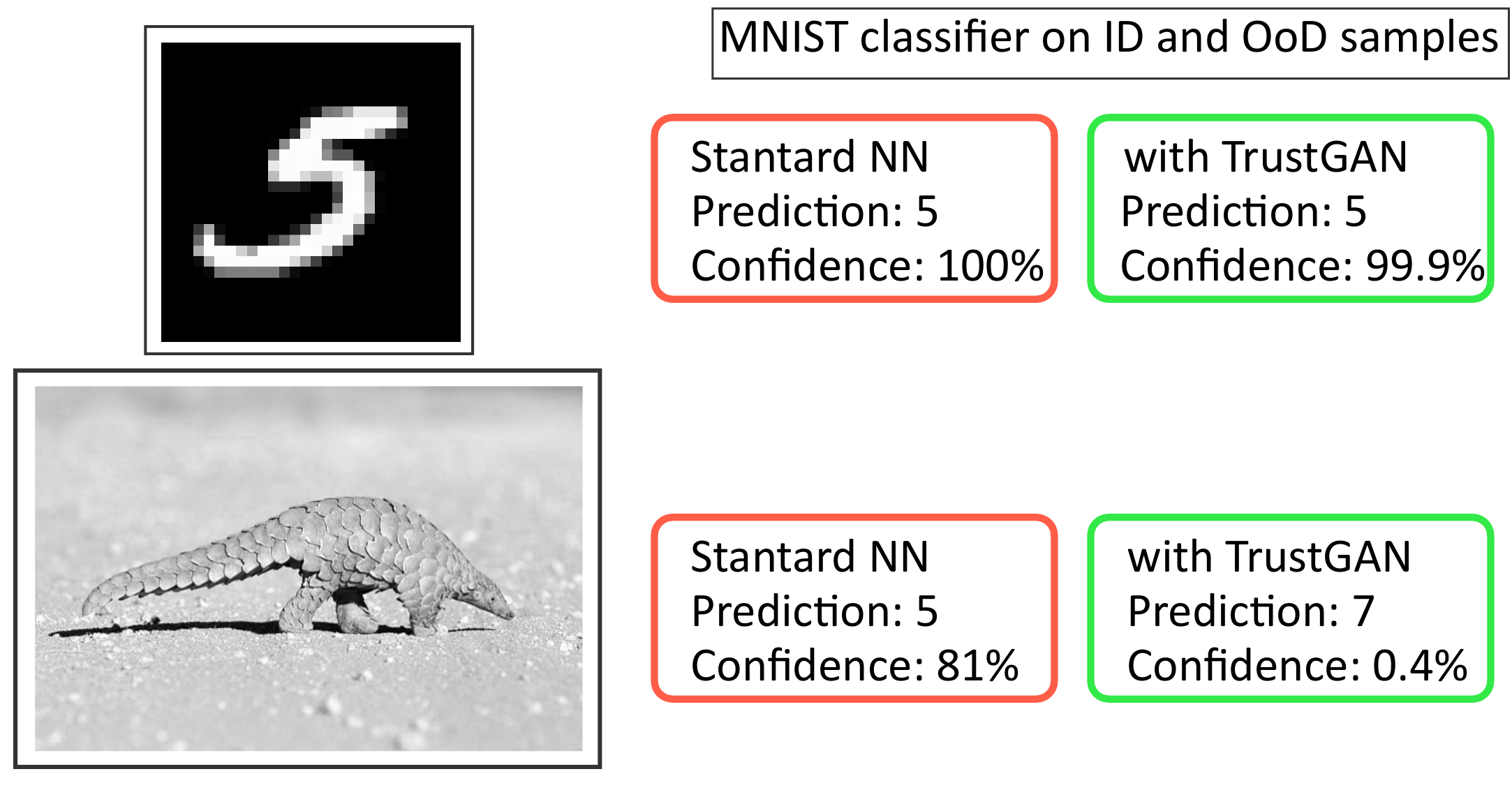}
    \caption{Example of in-distribution (ID) and out-of-distribution (OoD) sample
	images given to a number classifier trained on MNIST data.
	The results in red are for a model trained the standard way, and in green
	with TrustGAN.
	On an ID sample (top), both give the correct prediction with a large confidence, however
	on an OoD sample (bottom), a standard neural network predicts a 5 with 81\% confidence.
	This confidence drops to 0.4\%, when the neural network is
	trained with TrustGAN.}
    \label{figure::example_problem}
\end{figure}
One solution is to build a deep learning model which is as successful on a task as possible,
but which also is able to raise a flag or abstain if it is not confident enough.
Such a deep learning network returns now both a decision and an estimated confidence
on the decision. This confidence needs to be robust to rare samples in the training set
and more importantly to out-of-distribution (OoD) samples.
These samples are data unknown to the network,
e.g. if building a classifier to recognise helicopters from planes, a realistic
OoD sample could be the image of a bird.
Then a robust AI system has to return a low confidence for such an image.
Gathering OoD samples or worse OoD data sets is very tedious and most of
the time impossible, for similar reasons as the ones discussed above for
training data sets.

Standard training pipelines for deep learning models do not focus on the estimation
of the confidence on OoD.
Instead these pipelines focus on getting the best performances on the training
data set. That being said, most machine learning models still
output an estimation of the confidence on their decision, e.g. the
maximum class probability (MCP).
The estimation is known to be unreliable and overestimated
(see for example \cite{DBLP:journals/corr/abs-1910-04851}).
We show an example of such a flaw in figure~\ref{figure::example_problem},
where a number classifier, efficient at its task, robustly classifies the image
of a pangolin as the digit 5 (red boxes).

We present here TrustGAN, a generative adversarial network (GAN \cite{goodfellow2014generative})
pipeline targeting trustness and confidence.
The goal of this pipeline is to attack the confidence estimated by a target model
in order to improve upon it.
We present the effect TrustGAN can have on in-distribution (ID) and OoD samples
in figure~\ref{figure::example_problem} (green boxes).
The idea of the pipeline starts with the understanding that since OoD samples
are hard or impossible to gather and train on,
then we could leave a GAN learning to produce them.
Through these generated adversarial samples, the target network would learn both to be efficient
at its target task and to understand what ID samples look like.

TrustGAN is composed of two neural networks:
\begin{enumerate}
	\item a ``target model'' we want to be both as good as can be on a given
	task on a training data set, and be able to produce reliable confidence
	score on ID and OoD samples;
	\item a ``confidence-attacker'' network, i.e. a GAN, whose goal is to
	attack the confidence of the target network.
\end{enumerate}

We publicly release the code at github/ThalesGroup/trustGAN\footnote{\url{https://github.com/ThalesGroup/trustGAN}}.

Other works target at improving the estimated confidence a neural
network can have on its predictions.
TrustGAN can either replace some or work jointly
in a confidence improvement pipeline.
Among the literature, \cite{2015arXiv150602142G} uses dropout layers
as confidence estimations.
\cite{DBLP:journals/corr/abs-1910-04851} for example trains an extra
neural network to output confidence scores by learning to tell when
the target neural network fails on the training data set.
Other works like \cite{2017arXiv171109325L} uses a GAN
to learn a discriminator as a confidence estimator.
Their work is similar to ours, but implies that two different networks
have to be deployed on the target hardware: the target network
for the prediction and the discriminator for the estimated confidence.

To demonstrate how TrustGAN works and its performances, we first present in
section~\ref{section::adversarial_network_for_improved_confidence}
the training pipeline:
the specificities of the neural networks, of the different training losses,
and of the training process.
Then we present in section~\ref{application_to_classification_tasks}
the application of TrustGAN to the classification of numbers
based on their images and the classification of radio signal
modulations based on raw data.
Finally we conclude and draw perspectives in section~\ref{conclusion}.

\section{Adversarial network for improved confidence}
\label{section::adversarial_network_for_improved_confidence}

We present in this section the TrustGAN pipeline, that aims at training a given
neural network while improving its estimated confidence on in-distribution
and out-of-distribution samples.

\subsection{A target model and its confidence-attacker}

\begin{figure}[ht]
    \centering
    \includegraphics[width=0.99\linewidth]{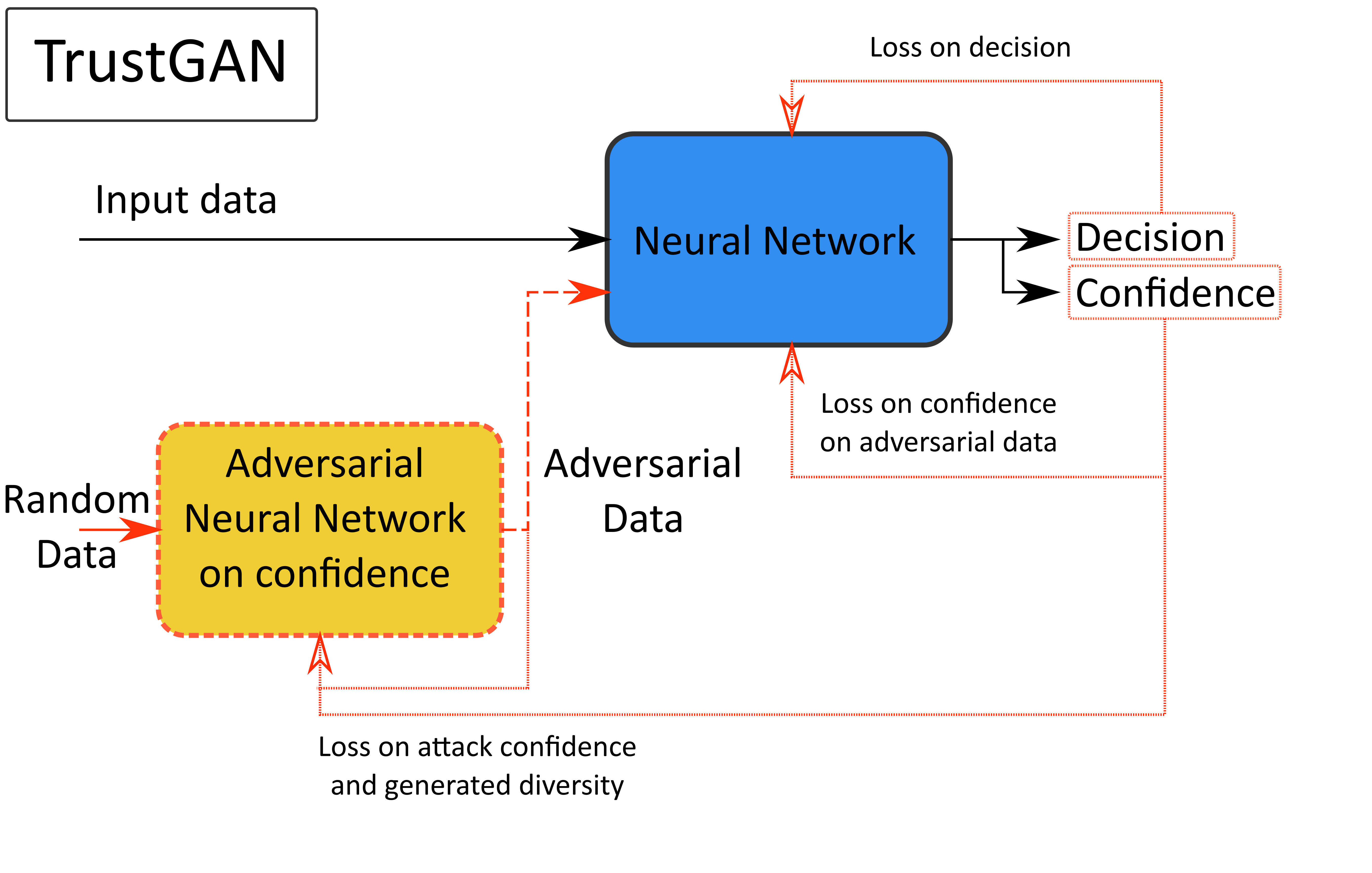}
    \caption{Presentation of the training process for TrustGAN.
	A target neural network in blue is improved through an adversarial network
	in yellow targeting its confidence. In this diagram, red lines show inputs used only
	during training, black lines are both for training and inference.}
    \label{figure::schema}
\end{figure}

The TrustGAN pipeline is presented in figure~\ref{figure::schema}.
It improves a neural network model, the target, that outputs a decision, e.g. a classification,
and a confidence on the given decision.
The estimated confidence on the decision can be defined in a variety of ways.
We work here in the most popular way, when working on a classification
task: the confidence is given by the maximum class probability
(MCP \cite{DBLP:journals/corr/HendrycksG16c}).
It is defined by:
\begin{equation}
	MCP = \max_{0 \leq i < n}{\hat{s_{i}}},
\end{equation}
where $n$ is the number of classes, $\hat{s_{i}}$ is the score (or probability)
for class $i$ estimated by the target network.
With this definition, MCP is between $1/n$, i.e. a random prediction
and $1$, i.e. $100\%$ confidence.
We define then the confidence by a re-normalization
of the estimated score $\hat{s_{i}}$
in order for the confidence to be a number between $0$ (not confident) and $1$ (very confident):
 \begin{equation}
	\hat{C_{i}} = \frac{\hat{s_{i}} - \frac{1}{n}}{1 - \frac{1}{n}}.
	\label{equation::conf_def}
\end{equation}
It is this confidence we aim at improving, while not hurting the predictive power of the
target model.

The target model is presented in blue in figure~\ref{figure::schema}: given an input data, it returns
a decision and a confidence.
The pipeline does not need to change the target model, however it must be provided
with a trainable white box implementation (e.g. Keras~\cite{chollet2015keras} or
PyTorch~\cite{paszke2019pytorch}) and its training metric (i.e. the training loss).
The fact that no changes have to be made on the target model, allows TrustGAN
to be agnostic and thus adapt to a large variety of cases.
The model parameters will be updated, if
the model has been trained beforehand or simply learned from start otherwise.

Beside the target model, a generative adversarial network
(GAN: \cite{goodfellow2014generative})
aims at fooling the target model into being confident on arbitrary inputs.
This confident attacker (yellow box on figure~\ref{figure::schema}) generates attacks,
i.e. target inputs, from random data.
This attacker must be tailored to the inputs the target model accepts. For example,
if the target model is a classifier running on images, the attacker must generates
from a vector of random numbers an image. However, if the target model is a classifier
running on 1D radio signals, the attacker must then generates 1D data.

In the configuration of figure~\ref{figure::schema}, no extra model is needed.
The role of the discriminator which tells if a sample is from the training distribution
or not, and which is standard in GANs,
is played by the target network itself, through a confidence loss. This simple
but special loss is described bellow (c.f. eqn.~\ref{equation::l_01}).
Other works, e.g. \cite{2017arXiv171109325L}, add a discriminator
to estimate the confidence.

In order to train both the target and the confidence-attacker model,
a training set and a validation set must be provided.
Furthermore, as said above, the target model loss must also be provided.
We detail bellow the training process, given in the black and red lines
of figure~\ref{figure::schema}.

\subsection{Training the target model}

The objective of the target model is two folds:
\begin{enumerate}
	\item be skilled at the task it has been designed for.
	This task is defined by the loss provided alongside the target model definition,
	and is estimated on the training set;
	\item refuse to be confident on data where it can not be, for example
	samples under represented in the training set or on out-of-distribution
	samples. These samples are represented by the data generated 
	by the confidence-attacker.
\end{enumerate}
These two goals can compete.
For example one method to succeed in the second task
is to never be confident on any decisions, and thus never provide a decision.
This goes against the goal of the first task.
The training pipeline has thus to find a balance between these
two goals.

For task number 1, the training loss $L_{00}$ is given by the provided loss
on the task the target network is aiming at, for example a classification model
could be trained with the cross-entropy loss.
This task is presented on figure~\ref{figure::schema} by "Loss on decision".

For task number 2, the training loss aims at setting the confidence to what it would
be for a random decision if the input data are not from the training set.
For a classification task, with $n$ classes, a random decision
is when the target network returns all $n$ probabilities to $1/n$. Thus the loss
for such a task is given by the soft-cross-entropy loss:
\begin{equation}
L_{01} = \frac{1}{\log{n}} \frac{1}{n} \sum_{i=0}^{n-1} \left[ - y_{i} \log{ \hat{s_{i}} } \right].
\label{equation::l_01}
\end{equation}
In this equation, $y_{i}$ is the target, here $y_{i} = 1/n$ and $\hat{s_{i}}$ is the estimated probability
of the target network, i.e. after softmax of the output logits $\hat{l_{i}}$. In this equation, $\log{n}$
allows to normalize the cross-entropy loss.

\subsection{Training the confidence-attacker}

From a vector of random numbers as inputs,
the confidence-attacker model must generate adversarial
samples as inputs to the target model with three objectives.
These objectives are:
\begin{enumerate}
	\item from random numbers generate data to
	fool the target network into being confident on its decision,
	whatever the decision is;
	\item from different random numbers generate different samples;
	\item from different random numbers induce different decisions of the target network.
\end{enumerate}
The first task aims at improving the estimated confidence by attacking it,
the other two are diversity tasks, aiming at attacking different aspects of
the target model.
These different tasks do not compete and can be obtained easily together.
However, they do compete with the two objectives of the target network.
The training pipeline has thus to find a balance between these
different goals.

For task number 1, the training loss is obtained by finding the class where
the confidence is maximum:
\begin{align}
L_{10} &= - \frac{1}{\log{n}}
	\max_{0 \leq i < n} \log{ \hat{s_{i}} }\\
& = \frac{1}{\log{n}} \times \left(
	-\max_{0 \leq i < n}{ (\hat{l_{i}})}
	+
	\log{\sum_{i=0}^{n-1} \exp{\hat{l_{i}}} }
	\right).
\end{align}
This training loss is thus pushing the confidence-attacker at generating
samples where the target model confidence is $C_{i} = 1$ at one of the classes
and $0$ elsewhere.

For task number 2, the goal is to generate diverse samples, if the random numbers
differ. The training loss is thus here defined by:
\begin{multline}
	L_{11} =
	\frac{1}{\sum_{j=0}^{N-1} \sum_{k=0}^{j-1} \overline{|r_{j} - r_{k}|^{m}} } \times\\
	\sum_{j=0}^{N-1} \sum_{k=0}^{j-1} \frac{
		\overline{|r_{j} - r_{k}|^{m}}
	}{
		1 + \overline{|a_{j} - a_{k}|^{m}}
	}.
\end{multline}
In this equation $j$ and $k$ are two different samples of a given set with $N$ samples.
This set can be for example a batch or a sub-sample of the batch, for faster computation.
For example, one can choose $N = n$.
The confidence-attacker model produces the adversarial sample $a_{j}$ from the random
numbers $r_{j}$ for the sample $j$.
The random numbers of the vector $r_{j}$ are sampled from the uniform distribution
$U\left( [0, 1) \right)$.
In this equation $m$ gives the order of the $m$-norm,
typically $m = 2$. This loss $L_{11}$ goes to $0$ if the generated samples are very different.

In a similar way, task number 3, compares the target model outputs:
\begin{multline}
	L_{12} =
	\frac{1}{\sum_{j=0}^{N-1} \sum_{k=0}^{j-1} \overline{|r_{j} - r_{k}|^{m}} } \times\\
	\sum_{j=0}^{N-1} \sum_{k=0}^{j-1} \frac{
		\overline{|r_{j} - r_{k}|^{m}}
	}{
		1 + \sum_{o=0}^{n-1} - \hat{s_{jo}} \log{ \hat{s_{ko}} }
	}.
\end{multline}
In this equation, $\hat{s_{jo}}$ is the predicted score by the target model
on the adversely generated sample $j$ of data $a_{j}$ for the class $o$.
This loss $L_{12}$ goes to $0$ if the generated scores are very different.

These three losses are combined into a single one:
\begin{equation}
	L_{13} = \frac{1}{3} \left( L_{10} + L_{11} + L_{12} \right),
\end{equation}
and are presented as "Loss on attack confidence and generated diversity"
in figure~\ref{figure::schema}.

\subsection{Training pipeline}

In order to obtain a target neural network, efficient at finding
a decision and estimating a robust confidence given a sample data, we must apply
the TrustGAN training pipeline. This pipeline is composed of three major steps per batch:
\begin{enumerate}
	\item train the confident-attacker model using random numbers with $L_{13}$;
	\item train the target model on adversely generated data with $L_{01}$;
	\item train the target model on data from the training set with $L_{00}$.
\end{enumerate}
These three steps are repeated for all epochs and for all batch in an epoch.
Different checkpoints are stored to follow the evolution of the different losses during training,
on both the training set and the validation set.
We store per epoch the state of the
best GAN of the epoch and the current state of the GAN. This allows to randomly repeat
previous attacks and prevents mode collapse, in a similar way as
experience replay in reinforcement learning (e.g. \cite{DBLP:journals/corr/abs-2007-06700}).
Furthermore, we store at each epoch, the current state of the target model.

Different parameters can be set to tune the relative power of the different losses and
attacks:
\begin{itemize}
	\item number of training epochs;
	\item batch size;
	\item number of epochs the target model is trained alone, i.e. only step 3, default value is $0$;
	\item number of step 1 for one step 3, default value is $1$;
	\item number of step 2 for one step 3, default value is $1$;
	\item random proportion of times step 1 and 2 are skipped, default value is $0\%$;
	\item random proportion of times adversarial samples are generated from a random
		GAN checkpoints instead of the current GAN state for step 2, default value is $10\%$.
\end{itemize}
Finally, we observe that it is best to design a simple GAN, with largely less parameters
than the target model. This allows to limit the available attacks the GAN can perform
on the target model, and thus do not prevent the target model from learning
the target task.

We store at each epoch different metrics: all the training losses.
They allow to track the learning process.
We also store at each epoch generated adversarial samples : the best attack of the
epoch and a random one at the end of the epoch.
They allow to visually understand, when possible, what the GAN is learning.
The available metrics can help choose the best target model, on the validation
set, given a compromise between the performances of the decisions and
the estimation of the confidence.
We here keep the best model according to the validation loss.

\subsection{Inference}

After TrustGAN training, the target model
is expected to give both good predictions and robust estimated confidence.
This model can thus be deployed as all standard models would be,
i.e. during inference we drop all red lines and boxes of figure~\ref{figure::schema}
and drop the confidence-attacker model.

Thanks to the re-normalization of the maximum class probability (eqn.~\ref{equation::conf_def}),
the confidence is provided between 0 (the decision is not confident) and 1 (the decision is confident).
This allows a user to now set a threshold $t$ on the confidence, and for a given sample $i$,
infer and get $(d_{i}, C_{i})$: the decision and its estimated confidence.
We now have:
\begin{itemize}
	\item if $C_{i} < t$ then the result is not confident enough, raise a flag or abstain from acting;
	\item if $C_{i} \geq t$ then the decision is $d_{i}$.
\end{itemize}
The threshold $t$ can be set by a user according to a balance between
rejections, i.e. needs a human decision, and automatic decisions by the neural network.

\section{Application to classification tasks}
\label{application_to_classification_tasks}

We test the TrustGAN training pipeline and compare it to a standard training pipeline in
two different applications: classification of numbers and classification of radio signals.
A standard training pipeline can be obtained easily from
section~\ref{section::adversarial_network_for_improved_confidence} by 
setting the number of epochs the target model is trained alone to be larger than
the actual number of epochs.
This allows to guarantee both standard and TrustGAN training are comparable.

\subsection{Classification of numbers}
\label{subsection::classification_of_numbers}

We apply TrustGAN training pipeline to the simple task of the classification of numbers given
an image. We use the MNIST (Modified National Institute of Standards and Technology) data set from \cite{deng2012mnist}. The task is thus a classification of $28$ pixels by $28$ pixels images with only one channel
onto $10$ classes. The training loss is thus a cross-entropy. The input images are min-max normalized over all channels, then rescaled to $[-1, 1]$. If all pixels have the same values, they are all set to $0$.
\begin{figure*}[ht]
    \centering
    \includegraphics[width=0.48\linewidth]{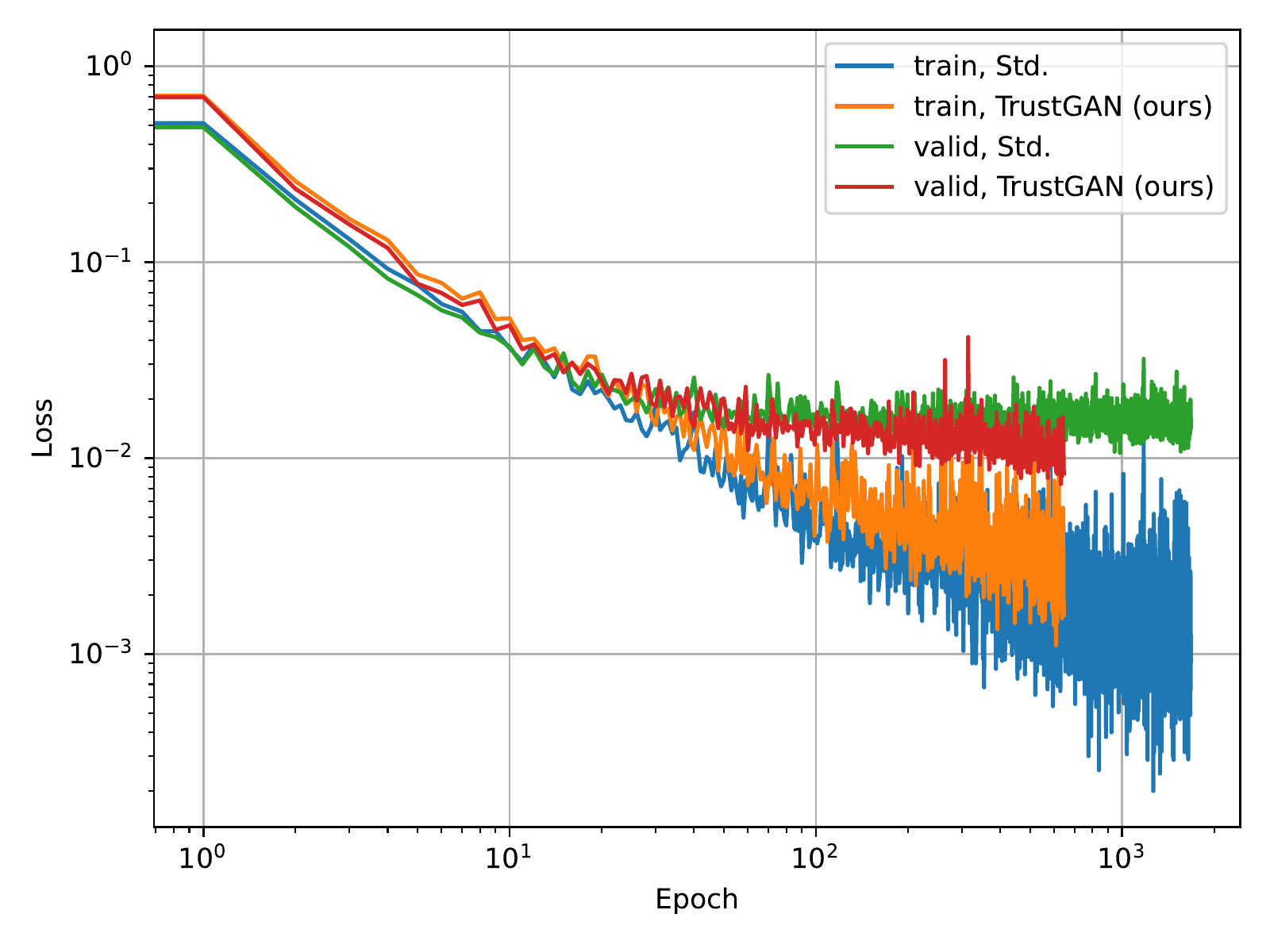}
    \includegraphics[width=0.48\linewidth]{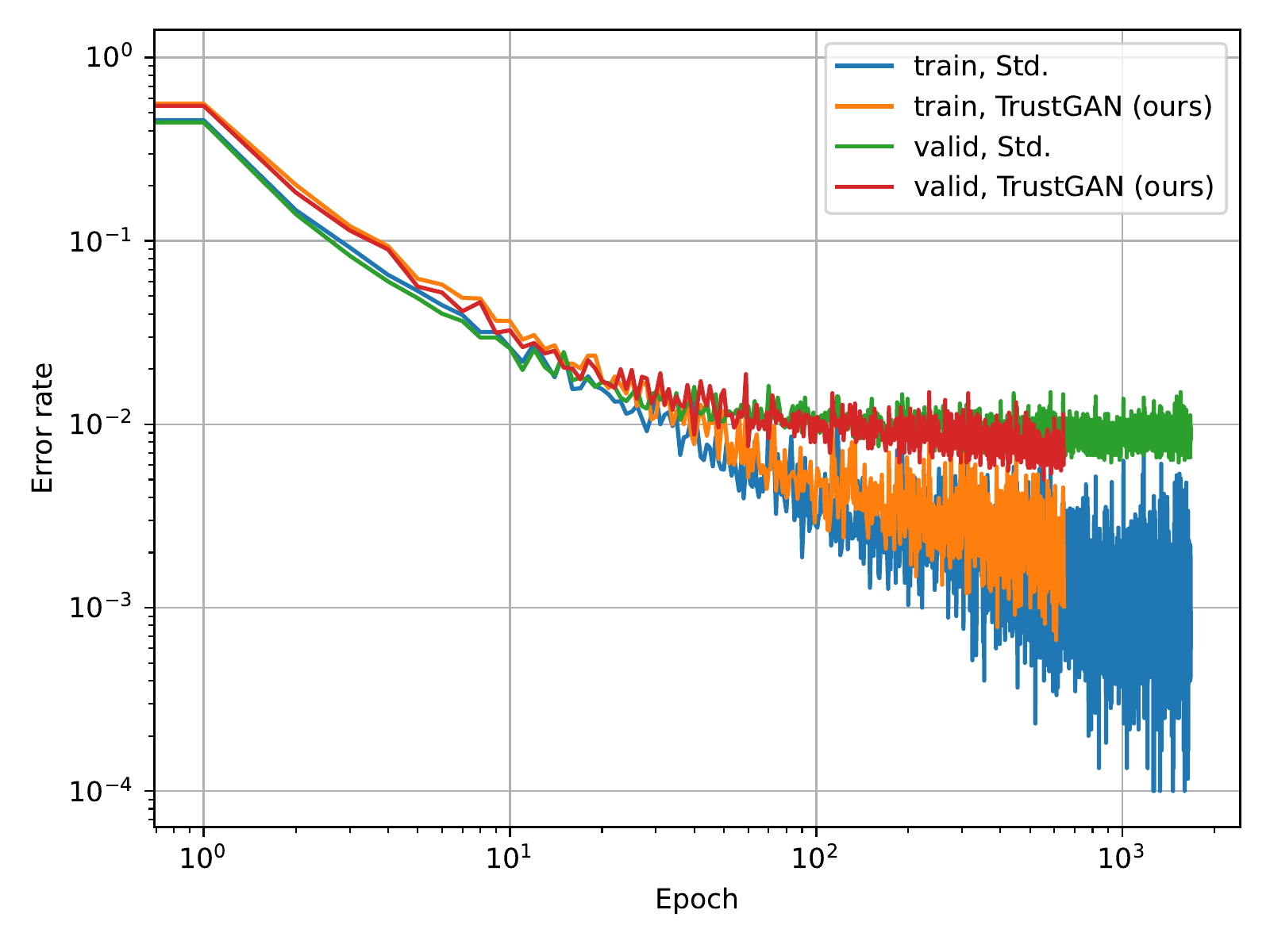}\\
    \caption{Target loss, i.e. $L_{00}$, (left) and error rate (right) training history for the training
	set and the validation set for a neural network trained
	to recognise numbers in MNIST data, either trained the standard way, or with TrustGAN.}
    \label{figure::training_history}
\end{figure*}
\begin{figure*}[ht]
    \centering
    \includegraphics[width=0.48\linewidth]{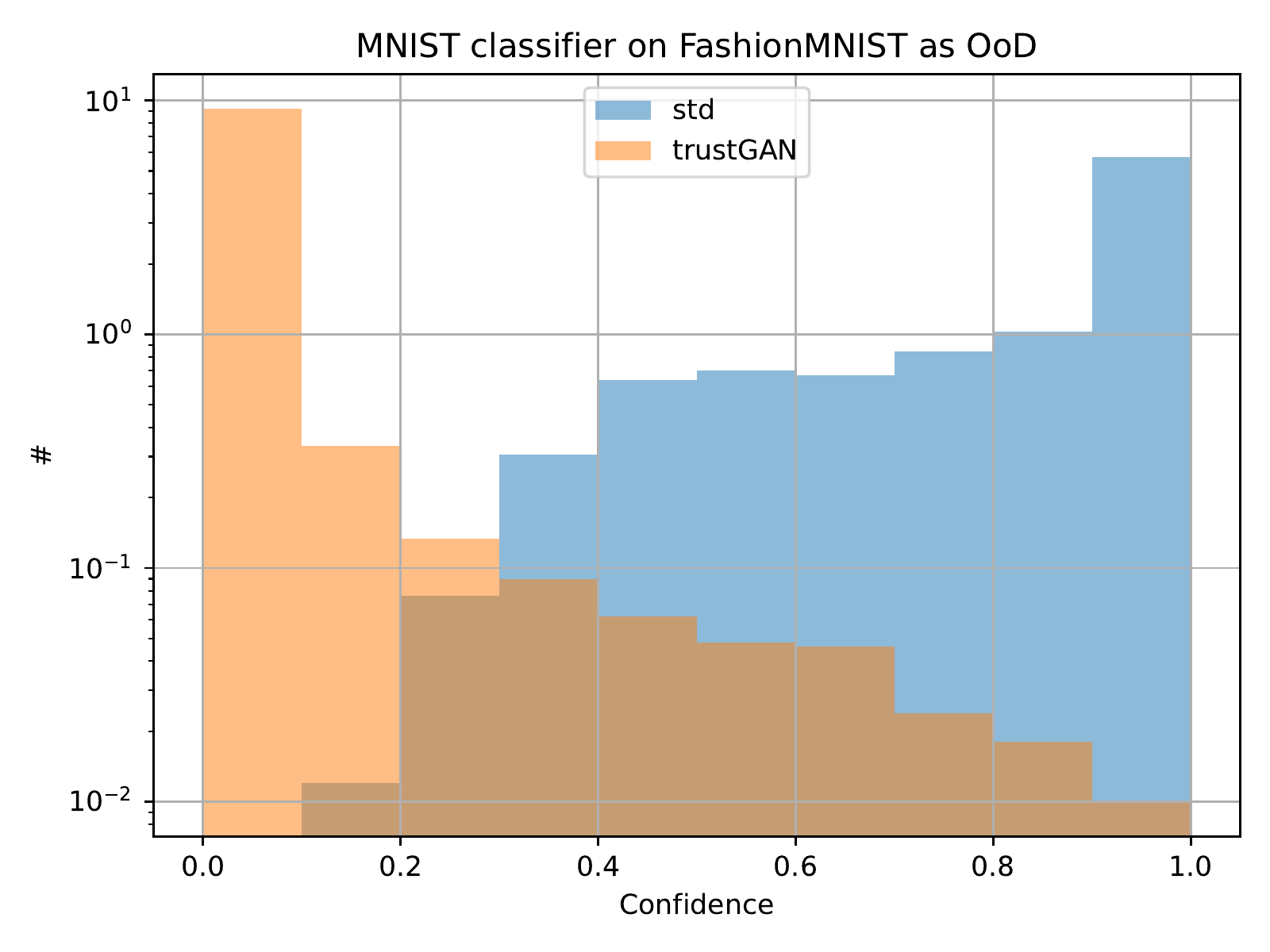}
    \includegraphics[width=0.48\linewidth]{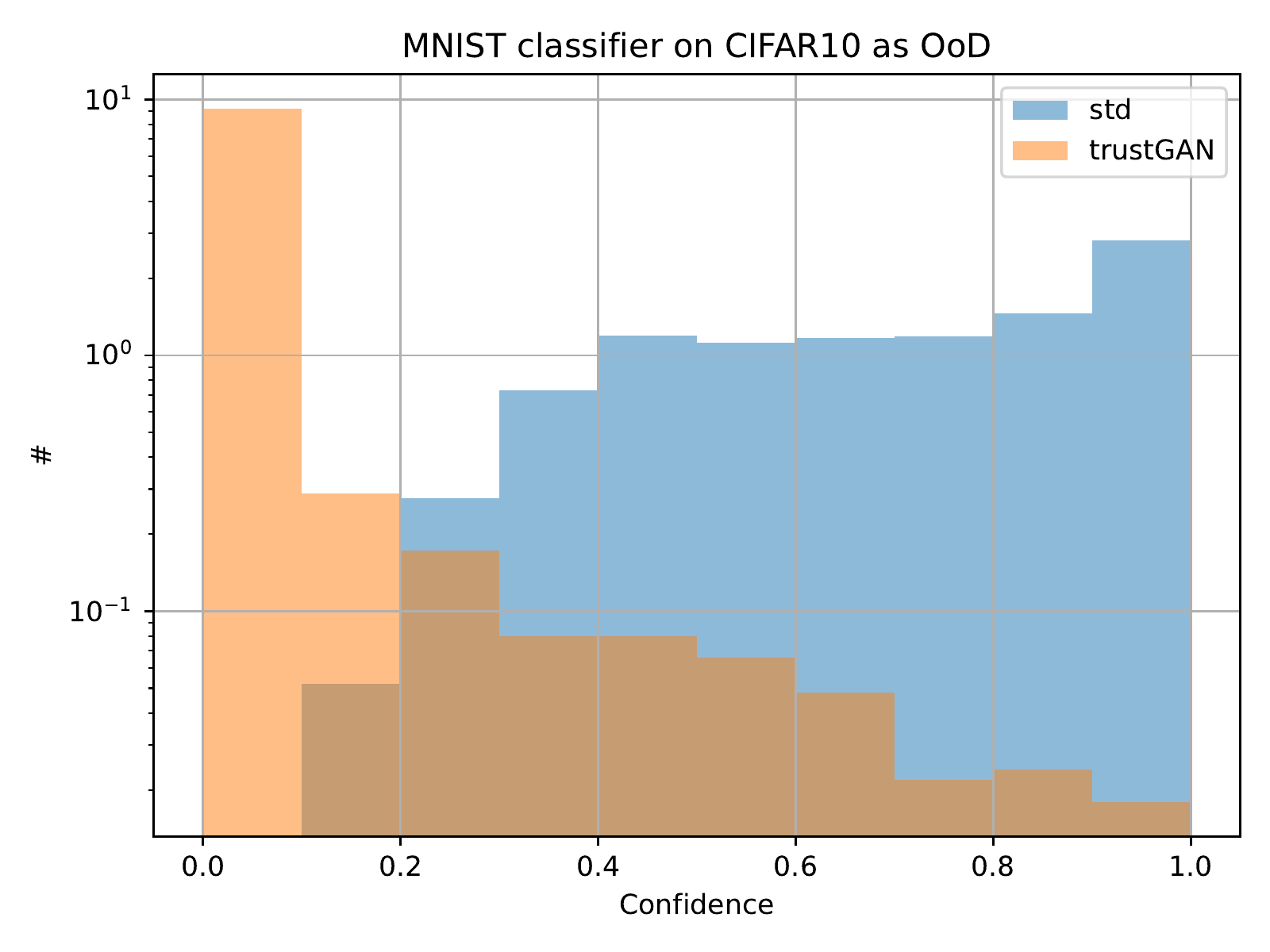}\\
    \caption{
	Distribution of the inferred confidence on two different out-of-distribution samples:
	FashionMNIST (left) and CIFAR10 (right). For both panels the neural network is trained
	to recognise numbers in MNIST data, in blue trained the standard way, in orange with TrustGAN.
	}
    \label{figure::classifier_mnist_ood_distrib_confidence}
\end{figure*}

Our target model is a 2D temporal convolutional neural network with residual connections
(see \cite{LeCun:1989:HDR:2969830.2969879}, \cite{DBLP:journals/corr/LeaVRH16} and \cite{7780459}) inspired by the architecture of \cite{9615851}.
Their structure is independent of the image length and width, and depends only on the number of channels.
This allows to test the architecture on random images of any sizes, for example from the internet (e.g. figure~\ref{figure::example_problem}).
It allows then to simply test how robust the network is against out-of-distribution samples.

The architecture of the confidence-attacker network is the same as that of the target model.
It only differs by few aspects. First, instead of a classification head, the network outputs images.
Second, the network is less deep than the target network.
This allows the GAN to attack the target model without preventing the latter from learning the target task.
Third, we use LeakyReLU (see \cite{Maas2013RectifierNI}), instead of ReLU (see \cite{agarap2018deep}),
which is known to be more adapted for GANs. Finally, for similar reasons we replace
weight norm layers (see \cite{DBLP:journals/corr/SalimansK16}) by batchnorm layers (see \cite{DBLP:journals/corr/IoffeS15}).
A tensor of random numbers of the same shape as the training set images is given to the model,
which then outputs through a hyperbolic tangent (tanh) activation layer a generated image.

The number of training epochs for the standard pipeline and for the TrustGAN
pipeline is set to an arbitrary high number (here $10~000$). Both pipelines are stopped when
no learning on the target loss is observed, on the validation set.
We periodically visually track the generated images to see if the GAN is learning useful
attacks.

Both the target loss and the target error rate learning history are shown on
figure~\ref{figure::training_history}.
We observe on both of the panels
that with the TrustGAN pipeline the target network takes more time to get
similar performances as the same model trained in the standard way.
This is expected, since with TrustGAN the target model is learning
the task while being attacked, which slows down learning.
However after some time, both models have similar performances in term
of both loss and error rate.
Appendix~\ref{appendix::gan_images} presents some randomly sampled GAN generated images.

In order to observe the resulting behavior of the estimated confidence, we use two
publicly available data sets as out-of-distribution samples:
FashionMNIST from \cite{DBLP:journals/corr/abs-1708-07747} and
CIFAR10  (Canadian Institute for Advanced Research) from \cite{cifar10}.
FashionMNIST is composed of black and white images of clothes,
and thus can be used without transformations.
CIFAR10 is composed of colored images of animals and vehicles;
we select then only the first color channel of each image.

A random selection of the images in both of the OoD data sets are inferred
by our MNIST classifiers. We thus get both a class and an estimated
confidence on this class.
Figure~\ref{figure::classifier_mnist_ood_distrib_confidence} presents the distribution
of the confidence for a model trained with a standard pipeline (blue) and
with the TrustGAN pipeline (orange).
The left panel presents
the results for FashionMNIST and the right panel for CIFAR10.
We observe in both cases that TrustGAN allows to push the confidence
of OoD to low values, i.e. both orange distributions are peaked at $0\%$
confidence. At the contrary, the standard model gives relatively high
confidences on these OoD samples.
This comparison shows how the TrustGAN pipeline helped the target
model not to be confident on OoD samples.

\subsection{Classification of radio signals}

The same experiment is conducted on 1D radio signals. We benefit from the publicly available
data set AugMod of \cite{9615851} to train a target model to recognise
radio signal modulations in raw 1D I/Q signals.
The target model of section~\ref{subsection::classification_of_numbers} is modified
to accept data with two channels, and 1D data. The latter is simply performed
by replacing 2D convolutions with 1D convolutions. We modify in the same
way the GAN. After both standard and TrustGAN training we obtain similar performances
in term of accuracy (c.f. table~\ref{table::quantitative_results}).

\begin{figure}[ht]
    \centering
    \includegraphics[width=0.99\linewidth]{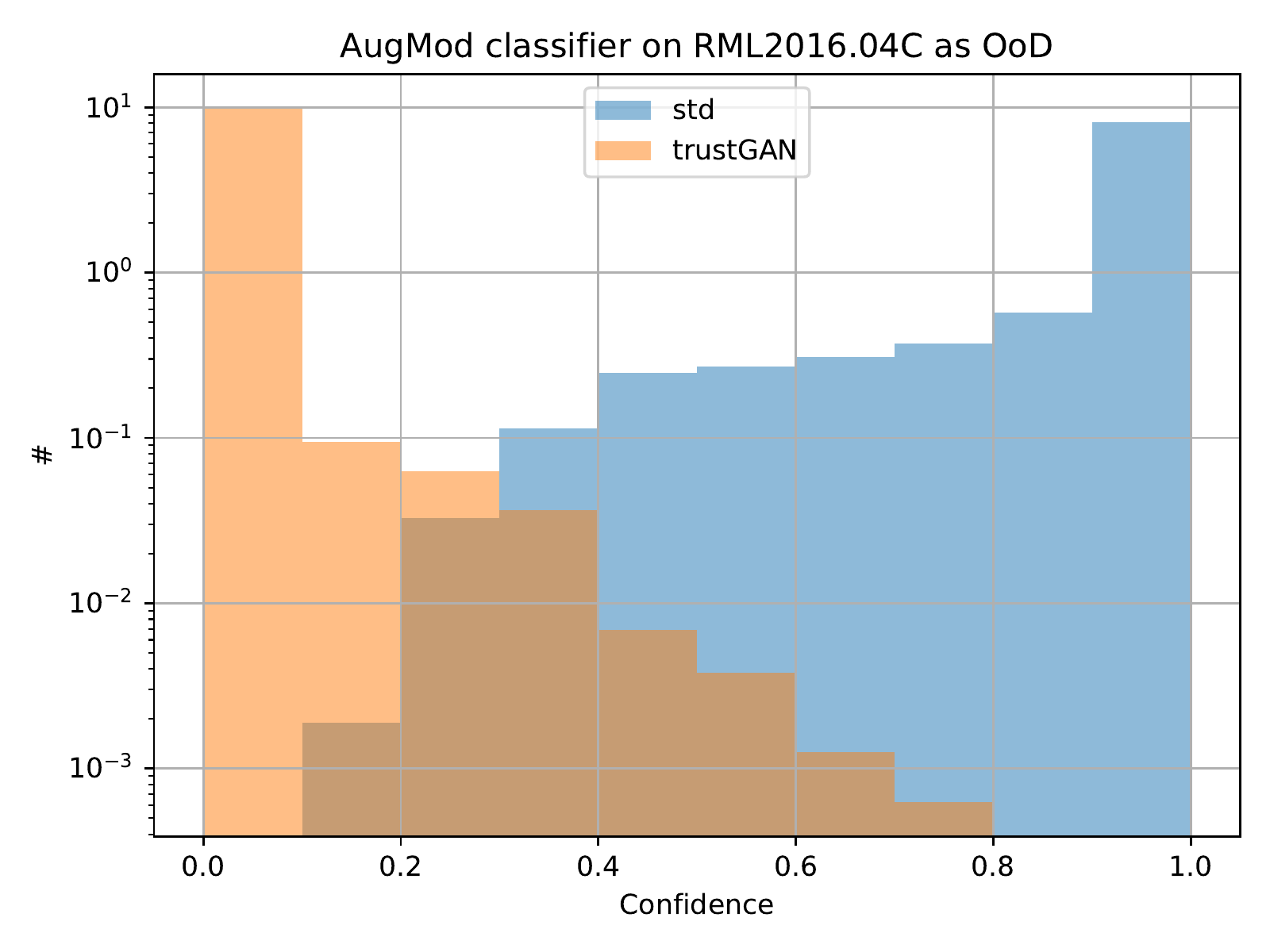}
    \caption{
	Distribution of the inferred confidence on out-of-distribution samples from
	RML2016.04C. The target neural network is trained
	to recognise modulations in AugMod data,
	in blue trained the standard way, in orange with TrustGAN.
	}
    \label{figure::classifier_augmod_ood_RML2016_04C_distrib_confidence}
\end{figure}

As for section~\ref{subsection::classification_of_numbers} we examine
the behavior of the estimated confidence with out-of-distribution samples.
The publicly available RML2016.04C data set from \cite{10.1007/978-3-319-44188-7_16}
is trimmed from modulations also present in the AugMod data set.
The resulting distribution of the inferred confidence is presented
on figure~\ref{figure::classifier_augmod_ood_RML2016_04C_distrib_confidence}.
We observe similarly that the target model is able to better estimate its confidence
on these out-of-distribution samples.

\subsection{Quantitative results}

Table~\ref{table::quantitative_results} presents quantitative results for the different tasks defined above:
\begin{itemize}
	\item MNIST-vs-FashionMNIST, where ID are MNIST and OoD are FashionMNIST;
	\item MNIST-vs-CIFAR10, where ID are MNIST and OoD are CIFAR10 first color channel;
	\item AugMod-vs-RML2016.04C, where ID are AugMod and OoD are RML2016.04C without modulations
		also present in AugMod.
\end{itemize}
\begin{table*}[ht]
	\caption{
		Performance of the estimated confidence using different metrics estimated on in-distribution (ID) and/or out-of-distribution (OoD) testset samples.
		For each metric, the best score is presented in bold.
	}
	\label{table::quantitative_results}
	\centering
	\scalebox{0.85}{
	\begin{tabular}{l | ll | lll | lll}
     &
	Accuracy                 $\uparrow$     &
	Loss                        $\downarrow$ &
	TPR@$\geq$0.90C   $\uparrow$     &
	FPR@$\geq$0.90C   $\downarrow$ &
	FPR@0.90TPR         $\downarrow$ &
	Confidence             $\downarrow$ &
	FPR@$\geq$0.90C   $\downarrow$ &
	FPR@0.90TPR         $\downarrow$  \\

     & (ID)  & (ID)  & (ID)  & (ID) & (ID) & (OoD) & (OoD) & (OoD vs. ID)   \\

     \noalign{\vskip 1mm} 
     \hline
     \hline
	\noalign{\vskip 2mm}

	MNIST-vs-FashionMNIST & & & & & \\
	{\hskip 1mm} Std., MCP                   & 0.99 & 0.034 & 0.98 & 0.0042 & 0.00019 & 0.84 & 0.57 & 0.15\\
     {\hskip 1mm} Std., MCDropout         &        &          & 0.97 & 0.0024 & 0.00020 & 0.76 & 0.41 & 0.13\\
	{\hskip 1mm} TrustGAN, MCP           & \textbf{0.99} &   \textbf{0.027}  &  \textbf{0.98} & 0.0032 & 0.00020 & 0.031 & 0.0010 & 0.0\\
     {\hskip 1mm} TrustGAN, MCDropout &       &           & 0.97 &  \textbf{0.0022} &  \textbf{0.0} &  \textbf{0.030} &  \textbf{0.00080} &  \textbf{0.0}\\
	 & & & & & \\

	MNIST-vs-CIFAR10 & & & & & \\
	{\hskip 1mm} Std., MCP                  &         &            &               &              &              &      0.71 & 0.28 & 0.015 \\
     {\hskip 1mm} Std., MCDropout        &          &            &               &             &              &      0.62 & 0.15 & 0.011 \\
	{\hskip 1mm} TrustGAN, MCP          &          &           &                &             &              &      0.031 & 0.0018 & 0.00020 \\
     {\hskip 1mm} TrustGAN, MCDropout &          &           &               &             &              &       \textbf{0.029} &  \textbf{0.0014} &  \textbf{0.0} \\
	 & & & & & \\

	\hline
	 & & & & & \\

	AugMod-vs-RML2016.04C & & & & & \\
	{\hskip 1mm} Std., MCP                   & 0.98 & 0.042 & \textbf{0.96} & 0.0039 & \textbf{0.00057} & 0.93 & 0.81 & 0.60 \\
     {\hskip 1mm} Std., MCDropout         &         &          & 0.92 & 0.0024 & 0.0018 & 0.82 & 0.50 & 0.46 \\
	{\hskip 1mm} TrustGAN, MCP           & \textbf{0.98} & \textbf{0.041} & 0.95 & 0.0027 & 0.00070 & 0.0060 & 0.0 & 0.0 \\
     {\hskip 1mm} TrustGAN, MCDropout  &         &          & 0.89 & \textbf{0.0015} & 0.0016 & \textbf{0.0057} & \textbf{0.0} & \textbf{0.0} \\

	\end{tabular}
	}
\end{table*}

We present results of the estimated MCP (maximum class probability, eqn. \ref{equation::conf_def})
for each task and for both a standard training pipeline (``Std., MCP'')
and the proposed pipeline (``TrustGAN, MCP'').
Along with MCP we present the results of Monte-Carlo Dropout (MCDropout, \cite{2015arXiv150602142G})
evaluated on both the standardly trained model (``Std., MCDropout'') and the TrustGAN model (``TrustGAN, MCDropout'').
MCDropout is performed with a mean of the softmax probabilities over $10$ realizations,
using the dropout layer defined in the target network, applied on the penultimate layer, with a rate of $0.3$.
The performances of the different technics are measured on different metrics estimated on the test sets:
\begin{itemize}
	\item accuracy (recall, true positive rate) of the ID test samples (higher the better);
	\item mean loss of the ID test samples (lower the better);
	\item TPR@$\geq$0.xxC, true positive rate at larger than 0.xx confidence (higher the better);
	\item $\mathrm{FPR}_{\mathrm{ID}}$@$\geq$0.xxC, false positive rate of in-distribution samples at larger than 0.xx confidence (lower the better);
	\item $\mathrm{FPR}_{\mathrm{ID}}$@0.xxTPR, false positive rate of ID samples at 0.xx true positive rate (lower the better);
	\item mean confidence of the OoD test samples (lower the better);
	\item $\mathrm{FPR}_{\mathrm{OoD}}$@$\geq$0.xxC, false positive rate of out-of-distribution samples at larger than 0.xx confidence (lower the better);
	\item $\mathrm{FPR}_{\mathrm{OoD}}$@0.xxTPR, false positive rate of OoD samples at 0.xx true positive rate (lower the better).
\end{itemize}

In the context of a classification task, the true positive rate (recall) is similar as the accuracy, defined as
a function of the confidence threshold $C$ on in-distribution samples (ID):
\begin{equation}
	TPR(C) = \frac{1}{N_{\mathrm{ID}}} \sum_{i \in \mathrm{ID}}
		\mathbb{1}_{ \hat{y_{i}} = y_{i}} \times \mathbb{1}_{ \hat{c_{i}} \geq C}.
\end{equation}
The false positive rate on ID samples
is defined as:
\begin{equation}
	{FPR}_{\mathrm{ID}}(C) = \frac{1}{N_{\mathrm{ID}}} \sum_{i \in \mathrm{ID}}
		\mathbb{1}_{ \hat{y_{i}} \neq y_{i}} \times \mathbb{1}_{ \hat{c_{i}} \geq C}.
\end{equation}
Finally, the false positive rate on out-of-distribution (OoD) samples
is defined as:
\begin{equation}
	{FPR}_{\mathrm{OoD}}(C) = \frac{1}{N_{\mathrm{OoD}}} \sum_{i \in \mathrm{OoD}}
		\mathbb{1}_{ \hat{c_{i}} \geq C}.
\end{equation}

We can observe from this table~\ref{table::quantitative_results} that on the ID testset the estimated metrics
are similar between the two training technics.
However, on OoD samples, TrustGAN produces significantly best scores.
We observe that MCDropout alone improves the performances for the standard models, however
either TrustGAN alone or TrustGAN combined with MCDropout are systematically better.
It is important to remember that all these results are observed even though the network has never seen these
specific OODs during training, but has only been presented GAN generated samples instead.

\section{Conclusion}
\label{conclusion}

In this study we presented a novel training pipeline, TrustGAN,
designed to robustify the estimated confidence a deep learning model
returns on a given prediction.
This pipeline allows to improve the behavior a given model has
with respect to out-of-distribution samples or samples rare enough in the
training set.

This TrustGAN pipeline does not need to change the target model in any
ways, except by updating its learned weights if the model is provided
already trained. The pipeline also requires a trainable white-box
implementation of the target model and of its training loss.
A generative adversarial network will learn to attack the target model.
These special attacks, targeting slowly the model confidence, result
in a model as predictive as a standard one, but also able to
tell when a given input is unknown or too complicated.

We test TrustGAN on two different tasks: the classification of numbers
using 2D images of the MNIST data set,
and the classification of the input radio modulation
in raw 1D signals of the AugMod data set.
On both of these tasks we observe no significant impact on the prediction accuracy,
but a large positive impact on the estimated confidence.

In the future we would like to better quantify the performances of the TrustGAN
pipeline in terms of the balance between missed prediction and out-of-distribution leakage
on more datasets.
Furthermore, we would like to compare the resulting confidence to other different
confidence estimators and robustifiers, e.g. data augmentation methods
and entropy methods.
We would then study how TrustGAN could be integrated to these estimators
and thus see how all could combine forces to improve
the estimated confidence and bring even safer, trustworthy, and deployable
deep learning models for critical systems.

\appendix

\label{appendix::gan_images}
\subsection{GAN generated images}

\begin{figure}[ht]
    \centering
    \includegraphics[width=0.99\linewidth]{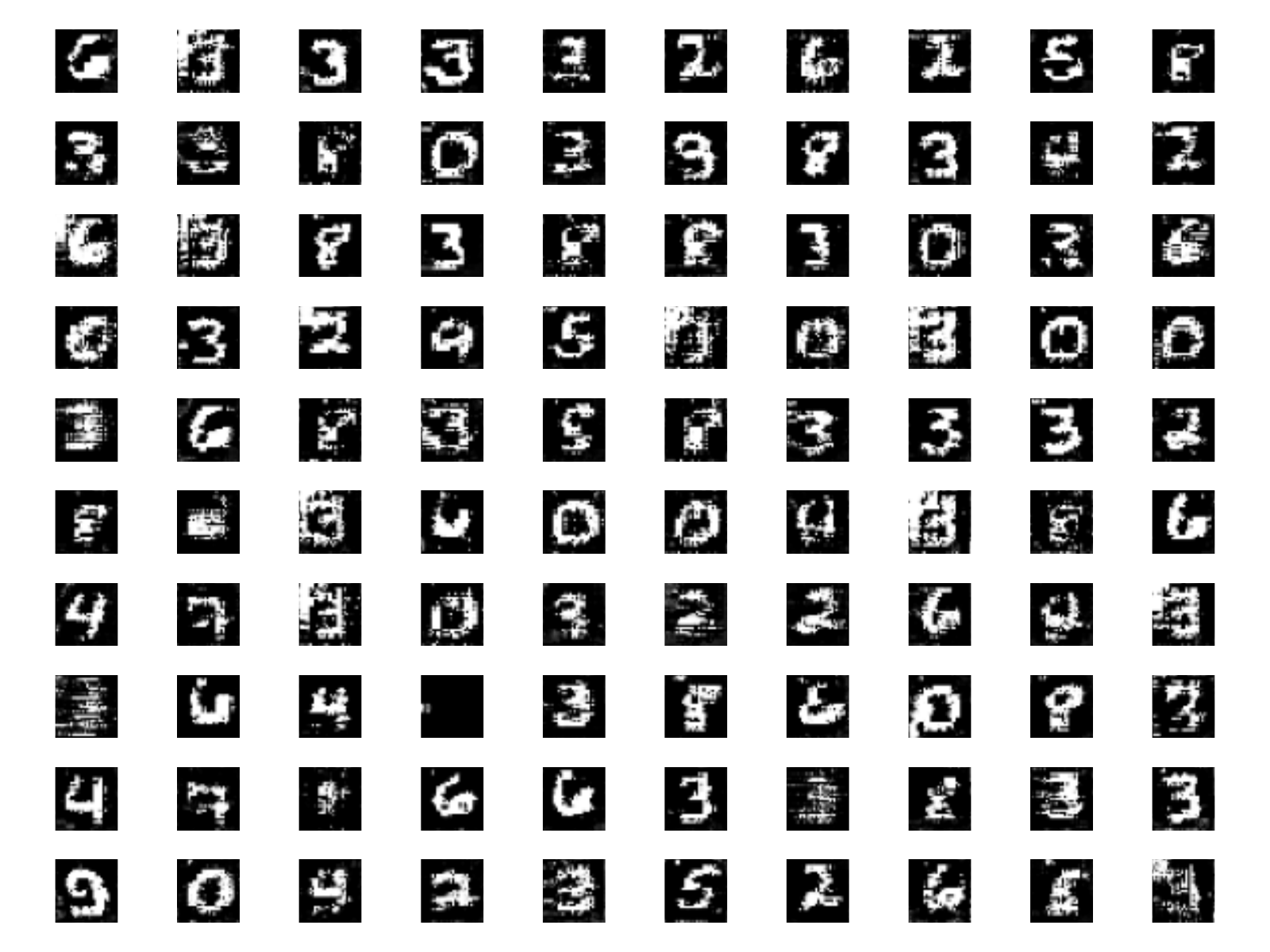}
    \caption{
	GAN generated images learned while
	training a classification neural network on the MNIST dataset.
	}
    \label{figure::example_gan_images}
\end{figure}

In figure~\ref{figure::example_gan_images}
we present some GAN generated images learned while
training a classification neural network on the MNIST dataset.
These images are randomly sampled after training is complete,
furthermore they are produced through randomly picking GANs
among all stored on discs and used for experience replays.

\bibliographystyle{IEEEtran}
\bibliography{conference_101719}

\end{document}